\documentclass[runningheads]{llncs}

\usepackage[T1]{fontenc}
\usepackage{graphicx}
\usepackage[ruled,vlined,linesnumbered]{algorithm2e}
\usepackage{algorithmic}
\usepackage{amsmath,amssymb}
\usepackage{amsfonts}
\usepackage{booktabs}
\usepackage{xcolor}
\usepackage{subcaption}
\usepackage{enumitem}
\usepackage[hyphens]{url}

\newcommand{\fref}[1]{Fig.~\ref{#1}}
\newcommand{\tref}[1]{Table~\ref{#1}}
\newcommand{\sref}[1]{Section~\ref{#1}}

\begin{document}

\title{What Helps---and What Hurts: Bidirectional Explanations for Vision Transformers}

\titlerunning{BiCAM}

\author{Qin Su \and
Tie Luo\thanks{Corresponding author.}}

\authorrunning{Su and Luo}

\institute{Department of Electrical and Computer Engineering\\
University of Kentucky, Lexington, KY, USA\\
\email{\{qin.su, t.luo\}@uky.edu}}

\maketitle
\begingroup
\renewcommand{\thefootnote}{}
\footnotetext{To appear in Proceedings of PAKDD 2026: {\it The $30^{th}$ Pacific-Asia Conference on Knowledge Discovery and Data Mining}.}
\endgroup

\begin{abstract}
Vision Transformers (ViTs) achieve strong performance in visual recognition, yet their decision-making remains difficult to interpret. We propose BiCAM, a bidirectional class activation mapping method that captures both supportive (positive) and suppressive (negative) contributions to model predictions. Unlike prior CAM-based approaches that discard negative signals, BiCAM preserves signed attributions to produce more complete and contrastive explanations. 
BiCAM further introduces a Positive-to-Negative Ratio (PNR) that summarizes attribution balance and enables lightweight detection of adversarial examples without retraining. 
Across ImageNet, VOC, and COCO, BiCAM improves localization and faithfulness while remaining computationally efficient. It generalizes to multiple ViT variants, including DeiT and Swin. These results suggest the importance of modeling both supportive and suppressive evidence for interpreting transformer-based vision models.

\keywords{Vision Transformer \and Explainable AI \and Adversarial Detection}
\end{abstract}

\section{Introduction}
Vision Transformers (ViTs)~\cite{dosovitskiy2020image} achieve state-of-the-art performance in classification~\cite{touvron2021training}, object detection~\cite{li2022exploring}, and segmentation~\cite{strudel2021segmenter}. Unlike convolutional neural networks (CNNs), ViTs leverage self-attention to model long-range dependencies without local inductive biases, enabling more expressive representations~\cite{naseer2021intriguing}, but their opaque decision-making hinders adoption in high-stakes domains. 

Existing ViT interpretability approaches include: (i) \textit{attention-based} methods~\cite{abnar2020} that multiply attention across layers and often over-smooth token differences; (ii) \textit{gradient-based} methods~\cite{LrpForViT,chen2022beyond,leem2024attention} that enhance attention with gradients but require full-network aggregation; and (iii) \textit{Shapley-based} methods~\cite{covert2023learning} that are computationally intensive.

\begin{figure}[t]
\centering
\includegraphics[trim=3mm 1mm 0 2mm, clip, width=0.8\linewidth]{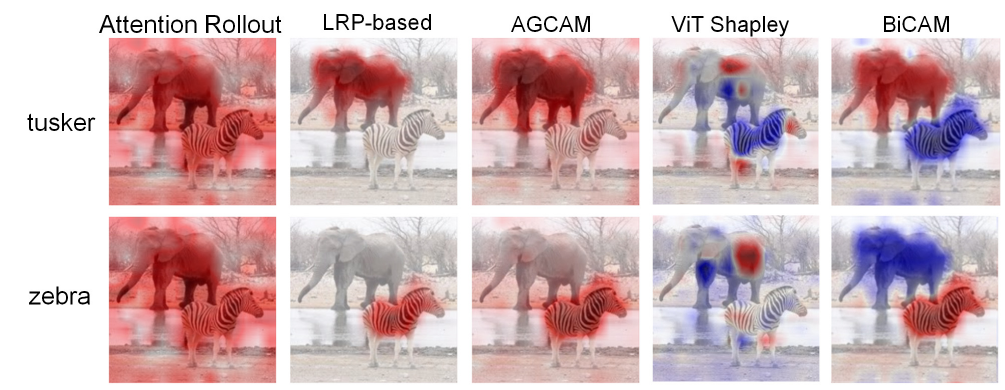}
\caption{BiCAM applied to ViT-B/16 on COCO. A query of ``tusker'' yields supportive (red) and suppressive (blue) attributions on the elephant and the zebra, respectively; yet querying ``zebra'' reverses the pattern. Baseline methods do not consistently produce this contrastive explanation.}
\label{fig:method_overview}
\end{figure}

We introduce \textbf{BiCAM}, a \textit{bidirectional class activation mapping} technique for ViTs. Unlike CAM-based methods that discard negative values, BiCAM treats them as informative, producing contrastive maps that explain both why the model predicts a class (\textit{supportive} evidence) and rejects alternatives (\textit{suppressive} evidence). This bidirectional view offers richer and more contrastive interpretations (\fref{fig:method_overview}), while requiring only a single forward--backward pass. 

BiCAM features a \textit{strategic layer aggregation} scheme. It selects only deeper transformer layers on the premise that class-discriminative information concentrates there~\cite{raghu2021vision}, augments attention in those layers with class-specific gradients, and fuses them with a simple summation in contrast to complex recursive operations~\cite{abnar2020} and weighted combinations~\cite{LrpForViT,chen2022beyond}, while empirically demonstrating improved effectiveness.

Beyond interpretability, we propose a simple {\em Positive-to-Negative Ratio} (PNR) that summarizes bidirectional attributions and enables effective detection of adversarial example attacks, including but not limited to PGD~\cite{madry2018towards}, C\&W~\cite{carlini2017towards}, and MI-FGSM~\cite{dong2018boosting}. 

BiCAM also generalizes to other ViT-based models with minimal modification, which we demonstrate using representative architectures such as DeiT~\cite{touvron2021training} and Swin~\cite{liu2021swin}. In summary, our main contributions are:
\begin{itemize}[topsep=0pt,partopsep=0pt,parsep=1pt]
    \item We propose BiCAM, a bidirectional attribution method that highlights supportive and suppressive factors with a single forward--backward pass, providing contrastive explanations for ViT-based models in single- and multi-object scenes.
    
    \item We introduce PNR, a simple metric derived from BiCAM to perform interpretable adversarial detection. The BiCAM design also integrates a principled layer aggregation strategy that focuses on deeper transformer layers where class-discriminative signals concentrate.
    
    \item  We empirically show strong localization, faithfulness, efficiency, and generalization performance of BiCAM. We also validate that PNR is a lightweight yet effective metric for adversarial detection.
\end{itemize}

\section{Related Work}

{\bf Raw Attention-based Methods.}
A common technique of ViT interpretability exploits transformer's native self-attention maps. Attention Rollout and Attention Flow \cite{abnar2020} recursively multiply attention matrices across layers to trace token influence, but often suffer from over-smoothing where tokens appear uniformly important. Prior work~\cite{serrano2019attention} shows that raw attention weights alone correlate poorly with feature importance.

{\bf Gradient-based and Relevance Propagation.}
To improve attribution fidelity, several methods combine attention with gradients. Beyond Attention~\cite{LrpForViT} adapts LRP~\cite{bach2015pixel} to ViTs, while Beyond Intuition~\cite{chen2022beyond} decomposes attribution into perception and reasoning. However, these methods aggregate across all layers without preserving signed contributions, limiting full-spectrum interpretability.

{\bf CAM-style Methods.}
 CAM~\cite{zhou2016learning} and Grad-CAM~\cite{GradCAM} have been adapted to ViTs by enhancing attention with gradients. AG-CAM \cite{leem2024attention} integrates class-specific gradients into attention heads; TS-CAM \cite{gao2021tscam} aligns tokens with class semantics for weakly supervised localization. CDAM~\cite{brocki2024class} modulates attention weights with class-specific gradients but computes relevance only at the final block, overlooking mid-layer cues.

{\bf Shapley-based Attribution.}
Inspired by SHAP~\cite{lundberg2017unified}, ViT-Shapley~\cite{covert2023learning} trains dataset-specific amortized explainers to estimate patch-level Shapley values, requiring substantial per-dataset training (19 hrs) with limited cross-dataset transferability. BiCAM requires only a single forward--backward pass with no training overhead.

In summary, existing methods predominantly focus on positive relevance or magnitude-based importance, overlooking the role of negative (suppressive) contributions. Also, unlike prior CAM variants, BiCAM differs not only in aggregation strategy but in preserving signed contributions throughout the pipeline, enabling contrastive interpretation and downstream signals (e.g., PNR) that are not accessible to magnitude-only methods.

\section{Methodology}\label{sec:method}

In a standard ViT, an input image is partitioned into patches, linearly embedded, and processed through Transformer blocks. Each block consists of multi-head self-attention (MSA) layers and feed-forward networks (FFNs). A [CLS] token aggregates global information for classification. BiCAM generates bidirectional attribution maps from this architecture. 

\subsection{Strategic Layer Aggregation}\label{sec:layer_agg}

Diverging from most existing ViT explainability methods \cite{abnar2020,LrpForViT,chen2022beyond,leem2024attention} that aggregate interpretability signals across all layers, BiCAM employs \textit{strategic layer aggregation} that selectively focuses on later transformer layers.

\textbf{Theoretical Basis.} Raghu et al.~\cite{raghu2021vision} shows that class-discriminative information concentrates in later layers, which is consistent with ViTDet~\cite{li2022exploring}. Thus, BiCAM selects the last $\ell$ blocks, and sets $\ell = 2L/3$ according to our ablation in \sref{sec:ablation}. This choice targets the layers where global semantic relationships are fully formed, while filtering out low-level structural noise.

\textbf{Layer Aggregation Strategy.} BiCAM computes and aggregates attributions from layers $L-\ell+1$ to $L$, where $L$ is the total number of layers. We extract, for each selected layer $l$, the attention maps $A^{(l)}$, value projections $V^{(l)}$, and gradients $\partial y_c/\partial o_{\text{cls}}^{(l)}$, and collect them ($A^{(l)}, V^{(l)}, \partial y_c/\partial o_{\text{cls}}^{(l)}$) to compute layer-wise attribution maps.
We then aggregate layer-wise masks via simple summation (detailed in \sref{sec:att}), rather than recursive matrix multiplication~\cite{abnar2020} or weighted combination~\cite{LrpForViT,chen2022beyond}. This preserves independent layer contributions without over-smoothing token differences or optimization overhead.

\subsection{BiCAM Attribution Mechanism}\label{sec:att}

Supportive and suppressive contributions are defined relative to the 
local gradient of the class logit with respect to token representations, combined with the token features themselves. Thus, negative values indicate directions that decrease the class score under infinitesimal perturbations, rather than strict counterfactual absence. BiCAM computes attributions through three steps:

\textbf{Step 1: Extract Attention and Values.}
For a ViT with $H$ heads and $N$ tokens, we extract from the last $\ell$ layers $l \in [L-\ell+1, L]$:
(i) Attention matrices $A^{(l)}_h \in \mathbb{R}^{B \times N \times N}$ for each head $h=1,...,H$, specifically CLS-to-patch attention $A^{(l)}_{h,0,:}$ via forward hooks before softmax.
(ii) Value projections $V^{(l)}_h \in \mathbb{R}^{B \times H \times N \times d_h}$ where $d_h = d/H$.

We apply softmax with temperature scaling~\cite{hinton2015distilling} to control the entropy of attention distributions, so we can soften attention to improve attribution stability:
\begin{equation}
\alpha^{(l)}_h = \text{softmax}(\frac{A^{(l)}_{h,0,:}}{T}), \quad h \in [1,H], l \in [L-\ell+1, L]
\end{equation}

\textbf{Step 2: Compute Gradients.}
We backpropagate the class-specific score $y_c$ with respect to [CLS] token embeddings:
\begin{equation}
w_c^{(l)} = \frac{\partial y_c}{\partial o^{(l)}_{\text{cls}}}, \quad l \in [L-\ell+1, L]
\end{equation}
where $o^{(l)}_{\text{cls}} = \text{concat}(o^{(l)}_{\text{cls},1}, ..., o^{(l)}_{\text{cls},H})$ is the concatenated [CLS] output across heads, with $o^{(l)}_{\text{cls},h} = \text{softmax}(A^{(l)}_{h,0,:}) \cdot V^{(l)}_h$.

\textbf{Step 3: Construct Attribution Maps.}
We aggregate attention, values, and gradients to form layer-wise masks:
\begin{equation}
\text{mask}^{(l)} = \sum_{h=1}^{H} \left((V^{(l)}_h \cdot w_c^{(l)}) \odot \alpha^{(l)}_h\right)
\end{equation}
where $\odot$ denotes element-wise multiplication. BiCAM transforms values through gradient weighting as in Grad-CAM~\cite{GradCAM}, with attention modulating these features. Layer-wise masks are summed:
$\text{mask} = \sum_{l \in [L-\ell+1, L]} \text{mask}^{(l)}$.
The resulting $[B, 1, N]$ mask is reshaped to $\sqrt{N{-}1} \times \sqrt{N{-}1}$ after discarding [CLS]. 

Importantly, no ReLU or clipping is applied throughout the computation, ensuring that both positive and negative contributions are preserved. The pseudocode is provided in Algorithm \ref{alg:bicam}.

\begin{algorithm}[t]
\footnotesize
\DontPrintSemicolon
\SetKwComment{Comment}{$\triangleright$ }{}
\caption{BiCAM Pseudocode} \label{alg:bicam}
\KwIn{ViT model $f$ with $L$ layers, aggregation window size $\ell$, image $x$ of shape ($H_t,W_d$), target class $c$}
\KwOut{Bidirectional attribution heatmap}
Register hooks to extract attention and gradients from layers $[L-\ell+1, L]$ \\
$y_c \gets f(x)[c]$\;
\For(\Comment*[f]{\scriptsize Forward pass}){$l \gets L - \ell + 1$ \KwTo $L$} {
    $\alpha^{(l)} \gets \text{softmax}(A^{(l)}_{h,0,:}/T)$ \Comment*{\scriptsize softmax with temperature-normed attention}
    $o_{\text{cls}}^{(l)} \gets \text{softmax}(A^{(l)}_{:,0,:}) \cdot V^{(l)}$ \Comment*{\scriptsize [CLS] output}
}
$y_c.\text{backward}()$\;
\For(\Comment*[f]{\scriptsize Backward pass}){$l \gets L$ down to $L-\ell+1$}{
    $w_c^{(l)} \gets \partial y_c/\partial o_{\text{cls}}^{(l)}$ \Comment*{\scriptsize Extract gradients}
    $\text{val\_grad}^{(l)} \gets V^{(l)} \cdot w_c^{(l)}$ \;
    $\text{mask}^{(l)} \gets \text{val\_grad}^{(l)} \odot \alpha^{(l)}$
}
$\text{mask}^{(l)} \gets \text{mask}^{(l)}.\text{sum}(\text{dim}=1)$ \Comment*{\scriptsize Aggregate heads}
$\text{mask} \gets \sum_{l} \text{mask}^{(l)}$ \Comment*{\scriptsize Aggregate layers}
$\text{mask} \gets \text{mask}[:,1:]$ \Comment*{\scriptsize Remove [CLS] token}
$\text{mask} \gets \text{reshape}(\text{mask}, (B, 1, \sqrt{N - 1}, \sqrt{N - 1}))$ \label{ln:reshape}\\
$\text{htmap} \gets \text{upsample}(\text{mask}, (B, 1, H_t, W_d))$ \\
\Return htmap\\
\end{algorithm}

\subsection{Positive-to-Negative Ratio (PNR)}\label{sec:pnr_def}

Our hypothesis is that clean samples tend to exhibit spatially structured positive–negative balance aligned with semantic regions, whereas adversarial perturbations \cite{tao2018attacks,yang2020ml}  introduce dispersed or exaggerated responses, skewing this balance. We thus propose a metric to quantify such distortion:
\begin{equation}
PNR = \frac{\sum_{i} ReLU(M_i)}{\sum_{i} ReLU(-M_i) + \epsilon}
\end{equation}
where $M_i$ is the attribution of ViT patch $i$ and $\epsilon$ prevents division by zero. A high PNR indicates dominance of supportive evidence while a low PNR suggests stronger suppressive influence. We then define \textbf{PNR Perturbation}: $\Delta PNR = PNR_{adv} - PNR_{clean}$, the difference between adversarial and clean samples.

\section{Evaluation}\label{sec:eval}\vspace{-1mm}
\subsection{Experimental Setup}\label{sec:setup}\vspace{-1mm}

We evaluate BiCAM on ViT-B/16~\cite{dosovitskiy2020image}, DeiT-B~\cite{touvron2021training}, and Swin-B~\cite{liu2021swin} at $224 \times 224$ resolution. Datasets include ImageNet \cite{russakovsky2015imagenet} for single-class attribution, VOC2012 \cite{everingham2010pascal} and COCO \cite{lin2014microsoft} for multi-label scenes, and adversarially perturbed examples for robustness.

We compare against \textbf{Attention Rollout}~\cite{abnar2020}, \textbf{LRP-based CAM}~\cite{LrpForViT}, \textbf{AGCAM}~\cite{leem2024attention}, and \textbf{ViT-Shapley}~\cite{covert2023learning}. For PNR, we use PGD~\cite{madry2018towards}, C\&W~\cite{carlini2017towards}, and MI-FGSM~\cite{dong2018boosting} attacks. 

Evaluation includes: (1) {\em Localization} via IoU, F1, Precision, Recall on ILSVRC, VOC, COCO annotations; (2) {\em Faithfulness} via iterative patch removal~\cite{covert2023learning}; (3) {\em PNR Perturbation} $\Delta PNR$.

\begin{figure}[t]
\centering\vspace{-3mm}
\includegraphics[width=0.9\columnwidth]{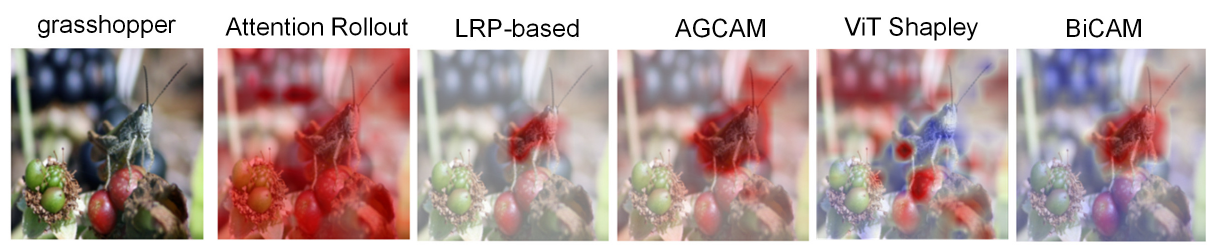}
\vspace{-3mm}\caption{Single-class attribution on ImageNet (grasshopper class; ViT-B/16). BiCAM highlights the target in red and background in blue.}
\label{fig:fine_grained}\medskip

\includegraphics[trim=3mm 2mm 0 2mm,clip,width=0.9\columnwidth]{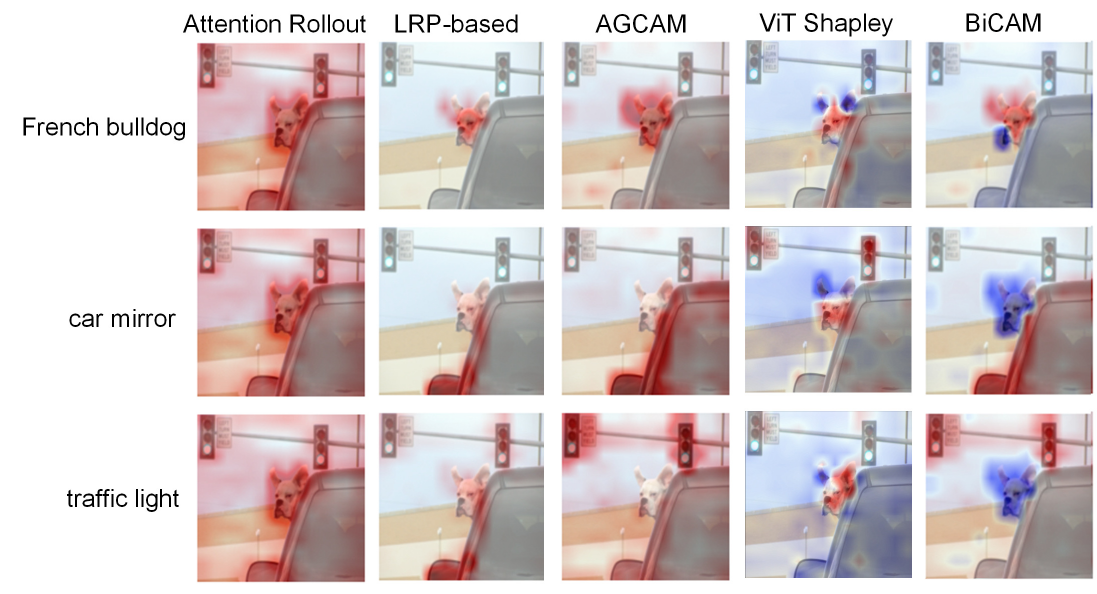}
\vspace{-3mm}\caption{Multi-class attribution on COCO (ViT-B/16). Each row queries a different class (bulldog, car mirror, traffic light). BiCAM consistently assigns supportive attribution (red) to the queried object and suppressive attribution (blue) to non-target objects.}
\label{fig:multi_class_scene}\vspace{-3mm}
\end{figure}

\vspace{-1mm}
\subsection{Qualitative Results}\label{sec:qualitative}

Our visualizations use red for supportive and blue for suppressive contributions. \fref{fig:fine_grained} shows single-class results from ImageNet, where BiCAM locates the target object and marks suppressive context. \fref{fig:multi_class_scene} shows multi-object results from COCO; BiCAM generates distinct attributions per queried class, consistently assigning supportive attribution to the target object and suppressive attribution to non-target objects.

\subsection{Localization Performance}\label{sec:localization}

Following~\cite{leem2024attention}, we normalize attribution maps to $[0,1]$ using min--max scaling and threshold the output at 0.5 to obtain binary masks. \tref{tab:localization} summarizes the results. On ImageNet, we report unified scores, whereas on VOC/COCO, positive and negative channels are evaluated separately against target and non-target annotations, respectively.

On ImageNet, BiCAM achieves the highest IoU (0.5419), F1 (0.6624), and recall (0.9288) among all methods. Its lower precision compared to AGCAM arises from preserving signed attributions, which leads to broader spatial coverage under unified evaluation. On VOC and COCO, where positive and negative channels are evaluated separately, this trade-off is no longer observed: BiCAM (Pos.) outperforms all baselines in IoU, F1, and precision. BiCAM (Neg.) has no direct counterpart since existing methods produce only positive attributions; moreover, vision models generally learn stronger representations for target objects than for background or non-target regions, placing BiCAM (Neg.) at an inherent disadvantage. Despite this, BiCAM (Neg.) achieves competitive performance relative to these baselines, suggesting that the suppressive maps capture semantically meaningful competing regions rather than random noise.

\begin{table}[ht]
\centering\vspace{-8mm}
\caption{Localization results on 3 datasets. ViT-Shapley is excluded because its amortized explainer could not scale to 1,000-class evaluation.}
\label{tab:localization}
\scriptsize
\setlength{\tabcolsep}{2pt}
\begin{tabular}{lccccc}
\toprule
\textbf{Method} & \textbf{Pix. Acc.} & \textbf{IoU} & \textbf{F1} & \textbf{Prec.} & \textbf{Rec.} \\
\midrule
\multicolumn{6}{l}{\textit{ImageNet-1k}} \\
Attn Rollout & 0.6209 & 0.3597 & 0.4893 & 0.7326 & 0.4657 \\
LRP-based & 0.5863 & 0.2029 & 0.3055 & \textbf{0.9110} & 0.2176 \\
AGCAM & \textbf{0.7341} & 0.5212 & 0.6515 & 0.8299 & 0.6276 \\
\textbf{BiCAM} & 0.6253 & \textbf{0.5419} & \textbf{0.6624} & 0.5901 & \textbf{0.9288} \\
\midrule
\multicolumn{6}{l}{\textit{VOC 2012}} \\
Attn Rollout & 0.8105 & 0.0686 & 0.1153 & 0.2604 & 0.1191 \\
LRP-based & 0.8420 & 0.1677 & 0.2464 & 0.4918 & 0.2021 \\
AGCAM & \textbf{0.8561} & 0.3561 & 0.4926 & 0.5932 & 0.5502 \\
\textbf{BiCAM (Pos.)} & 0.8559 & \textbf{0.3700} & \textbf{0.5104} & \textbf{0.6095} & \textbf{0.5863} \\
\textbf{BiCAM (Neg.)} & 0.7705 & 0.2588 & 0.3786 & 0.3642 & 0.5779 \\
\midrule
\multicolumn{6}{l}{\textit{COCO 2017}} \\
Attn Rollout & 0.8314 & 0.1332 & 0.2023 & 0.3276 & 0.2526 \\
LRP-based & 0.8523 & 0.1162 & 0.1871 & 0.4274 & 0.1835 \\
AGCAM & \textbf{0.8740} & 0.2807 & 0.3996 & 0.4210 & \textbf{0.6279} \\
\textbf{BiCAM (Pos.)} & 0.8707 & \textbf{0.2971} & \textbf{0.4191} & \textbf{0.5535} & 0.5154 \\
\textbf{BiCAM (Neg.)} & 0.8487 & 0.1141 & 0.1724 & 0.2711 & 0.2074 \\
\bottomrule
\end{tabular}\vspace{-10mm}
\end{table}

\subsection{Faithfulness via Feature Removal}\label{sec:faithfulness}

We evaluate attribution faithfulness via perturbation experiments~\cite{samek2016evaluating}, removing patches in descending order of importance \textbf{Most Important Feature (MIF)} or ascending order \textbf{Least Important Feature (LIF)} and define
$\text{Faithfulness} = \text{AUC}_{\text{LIF}} - \text{AUC}_{\text{MIF}}$. 

\begin{figure}[ht]
\centering
\begin{subfigure}[b]{0.47\columnwidth}
\includegraphics[width=\textwidth]{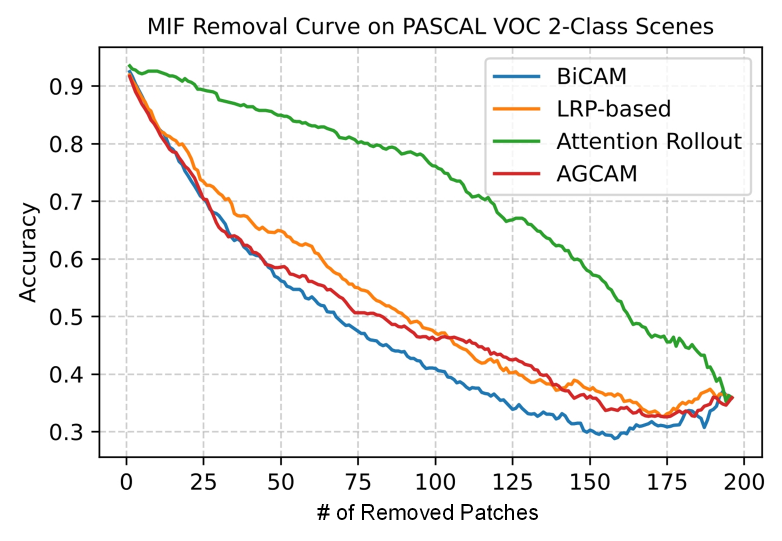}
\caption{VOC - MIF removal}
\end{subfigure}
\hfill
\begin{subfigure}[b]{0.47\columnwidth}
\includegraphics[width=\textwidth]{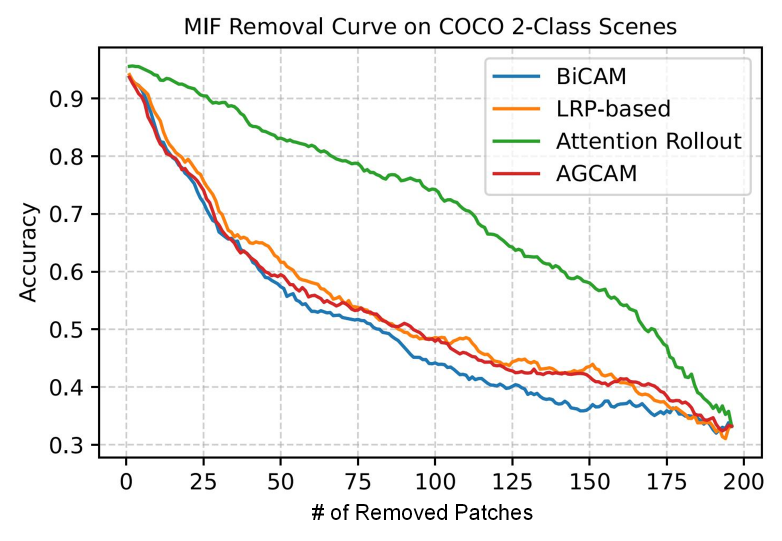}
\caption{COCO - MIF removal}
\end{subfigure}

\vspace{0.2em}

\begin{subfigure}[b]{0.47\columnwidth}
\includegraphics[width=\textwidth]{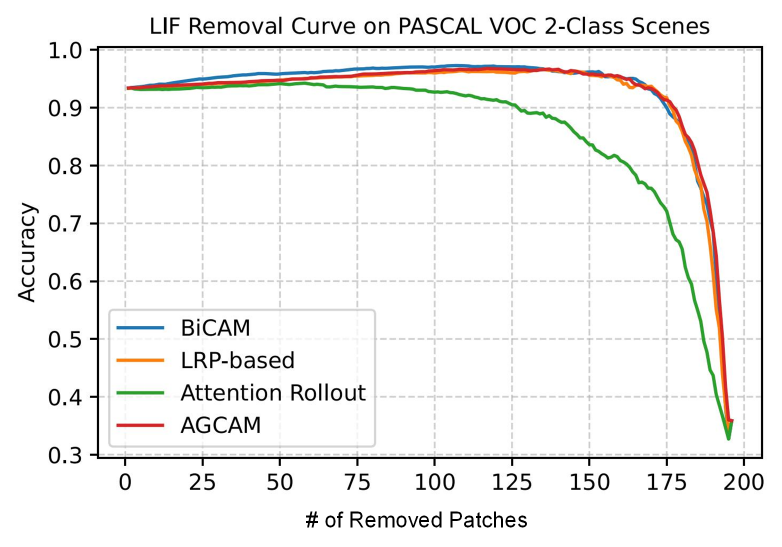}
\caption{VOC - LIF removal}
\end{subfigure}
\hfill
\begin{subfigure}[b]{0.47\columnwidth}
\includegraphics[width=\textwidth]{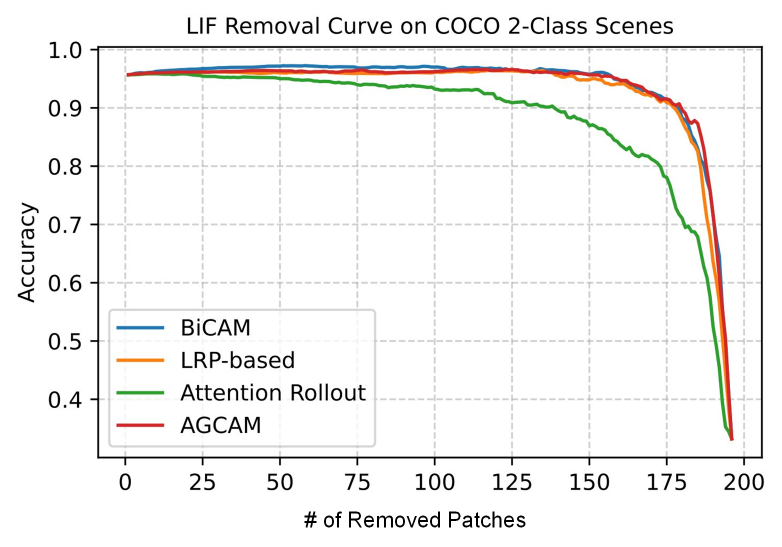}
\caption{COCO - LIF removal}
\end{subfigure}

\caption{Faithfulness evaluation via perturbation on VOC and COCO (ViT-B/16). Curves show the mean across five random seeds. Higher faithfulness is indicated by steeper accuracy degradation in MIF removal and flatter curves in LIF removal.}
\label{fig:perturbation}
\end{figure}

We remove the 196 patches one by one (each masked to zero), in the order of their attribution scores (ascending: LIF; descending: MIF). \tref{tab:perturbation} summarizes the faithfulness scores averaged over five random seeds, and \fref{fig:perturbation} shows the corresponding perturbation curves. Pairwise Wilcoxon signed-rank tests ($p < 0.001$) confirm that BiCAM significantly outperforms all baselines on all datasets, indicating better alignment between attribution importance and model behavior.

\begin{table}[t]
\centering
\caption{Faithfulness scores. ViT-Shapley is excluded because it achieves only 0.294 on the 10-class ImageNette.}
\label{tab:perturbation}
\renewcommand{\arraystretch}{0.95}
\scriptsize
\begin{tabular}{lcccc}
\toprule
& \textbf{Attn Rollout} & \textbf{LRP} & \textbf{AGCAM} & \textbf{BiCAM} \\
\midrule
\multicolumn{5}{l}{\textbf{ImageNet}} \\
LIF$\uparrow$ & 0.4739 & 0.5140 & 0.5298 & \textbf{0.5478} \\
MIF$\downarrow$ & 0.2053 & 0.1736 & \textbf{0.1607} & 0.1654 \\
Faith$\uparrow$ & 0.2685 & 0.3404 & 0.3691 & \textbf{0.3824} \\
\midrule
\multicolumn{5}{l}{\textbf{VOC 2-class}} \\
LIF$\uparrow$ & 0.8623 & 0.9231 & 0.9280 & \textbf{0.9313} \\
MIF$\downarrow$ & 0.7142 & 0.5201 & 0.5010 & \textbf{0.4686} \\
Faith$\uparrow$ & 0.1481 & 0.4030 & 0.4270 & \textbf{0.4626} \\
\midrule
\multicolumn{5}{l}{\textbf{COCO 2-class}} \\
LIF$\uparrow$ & 0.8856 & 0.9307 & 0.9376 & \textbf{0.9407} \\
MIF$\downarrow$ & 0.7052 & 0.5327 & 0.5235 & \textbf{0.4997} \\
Faith$\uparrow$ & 0.1805 & 0.3980 & 0.4141 & \textbf{0.4410} \\
\bottomrule
\end{tabular}
\end{table}

\vspace{-2mm}
\subsection{PNR Analysis}\label{sec:pnr}

Adversarial perturbations tend to disrupt model prediction more in multi-object scenes due to the multiple competing objects. Hence we choose VOC 2012 ($\geq$2 classes) for evaluation. In such scenes, we also find that BiCAM's negative attributions correctly fall on competing objects rather than diffuse background, leading to a stable denominator of PNR. \tref{tab:pnr_detection} shows that all the tested attacks inflate PNR$_{adv}$ and hence $\Delta PNR$ notably. Using $\Delta PNR$ values, together with a threshold we derive by maximizing {\em Youden's J statistic} (sensitivity $+$ specificity $- 1$), we achieve accurate detection indicated by the high AUROC and AUPR. This indicates that $\Delta PNR$ provides a simple, lightweight, yet effective adversarial detection signal without retraining.

\begin{table}[t]
\centering
\vspace{-3mm}\caption{PNR-based adversarial detection on VOC 2012 multi-object scenes.}
\label{tab:pnr_detection}
\scriptsize
\setlength{\tabcolsep}{2.5pt}
\begin{tabular}{lccccc}
\toprule
\textbf{Attack} & \textbf{$\Delta$PNR} & \textbf{std} & \textbf{AUROC} & \textbf{AUPR} & \textbf{Threshold} \\
\midrule
Clean & 0.00 & N/A & - & - & - \\
\midrule
PGD & +0.57 & 0.31 & 0.781 & 0.749 & 0.248 \\
C\&W & +0.79 & 0.37 & {0.842} & {0.808} & 0.332 \\
MI-FGSM & +0.46 & 0.27 & 0.764 & 0.731 & 0.219 \\
\midrule
Avg. & +0.61 & 0.32 & 0.796 & 0.763 & 0.266 \\
\bottomrule
\end{tabular}\vspace{-5mm}
\end{table}

\vspace{-2mm}
\subsection{Computational Efficiency}\label{sec:efficiency}\vspace{-1mm}

\tref{tab:runtime} reports inference costs (time and GPU memory) on ViT-B/16, measured on an RTX 4090 over 1,000 ImageNet images. BiCAM is 8.4$\times$ faster than LRP with no training overhead, and LRP requires large memory to store intermediate relevance scores at every layer. ViT-Shapley requires per-dataset retraining.

\begin{table}[!ht]
\centering
\vspace{-5mm}\caption{Runtime comparison (ViT-B/16, RTX 4090, ImageNet-1k).}
\label{tab:runtime}
\scriptsize
\setlength{\tabcolsep}{4pt}
\begin{tabular}{lccc}
\toprule
\textbf{Method} & \textbf{ms/img} & \textbf{MB/img} & \textbf{Train time} \\
\midrule
Attn Rollout & 23.3 & 0.15 & None \\
LRP & 134.6 & 97.2 & None \\
ViT-Shapley & 10.1$^*$ & -- & 19 hrs$^*$ \\
AGCAM & 20.0 & 0.22 & None \\
\textbf{BiCAM} & \textbf{16.0} & 0.24 & None \\
\bottomrule
\end{tabular}
\vspace{0.2em}

\footnotesize
$^*$ImageNette (10 classes), measured on RTX 2080 Ti~\cite{covert2023learning}.
\end{table}\vspace{-3mm}

\subsection{Ablation Studies}\label{sec:ablation}\vspace{-1mm}

We validate key design choices on PASCAL VOC 2012 with ViT-B/16. As shown in \tref{tab:ablation}, $\ell{=}2L/3$ achieves the best localization; using all layers ($\ell{=}L$) introduces early-layer noise (IoU drops 23\%), while too few layers ($\ell{=}L/3$) loses mid-layer signals. This supports the hypothesis that early layers capture low-level features that are less class-discriminative.

For temperature, we choose $T{=}2$ as it balances sharpness and smoothness the best, while $T{=}1$ produces overly peaky attributions (IoU\,=\,0.105) and $T{=}3$ over-smooths.

\begin{table}[!ht]
\centering
\caption{Ablation on layer window $\ell$ and temperature $T$ (ViT-B/16, VOC 2012).}
\label{tab:ablation}
\scriptsize
\setlength{\tabcolsep}{2.5pt}
\begin{tabular}{lcccc|ccc}
\toprule
& \multicolumn{4}{c|}{\textit{Layer window $\ell$ \;($T{=}2$)}} & \multicolumn{3}{c}{\textit{Temperature $T$ \;($\ell{=}8$)}} \\
& $\ell{=}12$ & \textbf{$\ell{=}8$} & $\ell{=}4$ & $\ell{=}1$ & $T{=}1$ & $\mathbf{T{=}2}$ & $T{=}3$ \\
\midrule
Pix.Acc & 0.8505 & \textbf{0.8559} & 0.8498 & 0.8413 & 0.8402 & \textbf{0.8559} & 0.8492 \\
IoU     & 0.2840 & \textbf{0.3700} & 0.2093 & 0.1491 & 0.1051 & \textbf{0.3700} & 0.2060 \\
F1      & 0.4043 & \textbf{0.5104} & 0.3173 & 0.2277 & 0.1711 & \textbf{0.5104} & 0.3140 \\
Prec.   & 0.5980 & \textbf{0.6095} & 0.6765 & 0.5343 & 0.6401 & 0.6095 & \textbf{0.6789} \\
Rec.    & 0.3949 & \textbf{0.5863} & 0.2649 & 0.1824 & 0.1144 & \textbf{0.5863} & 0.2589 \\
Faith.  & \textbf{0.4746} & 0.4626 & 0.4570 & 0.3499 & 0.4472 & 0.4626 & \textbf{0.4900} \\
\bottomrule
\end{tabular}
\end{table}

\subsection{Generalization to ViT Variants}\label{sec:generalization}

\begin{figure}[tb]
\centering\vspace{-3mm}
\includegraphics[trim=5mm 1mm 0 1mm,clip,width=0.75\columnwidth]{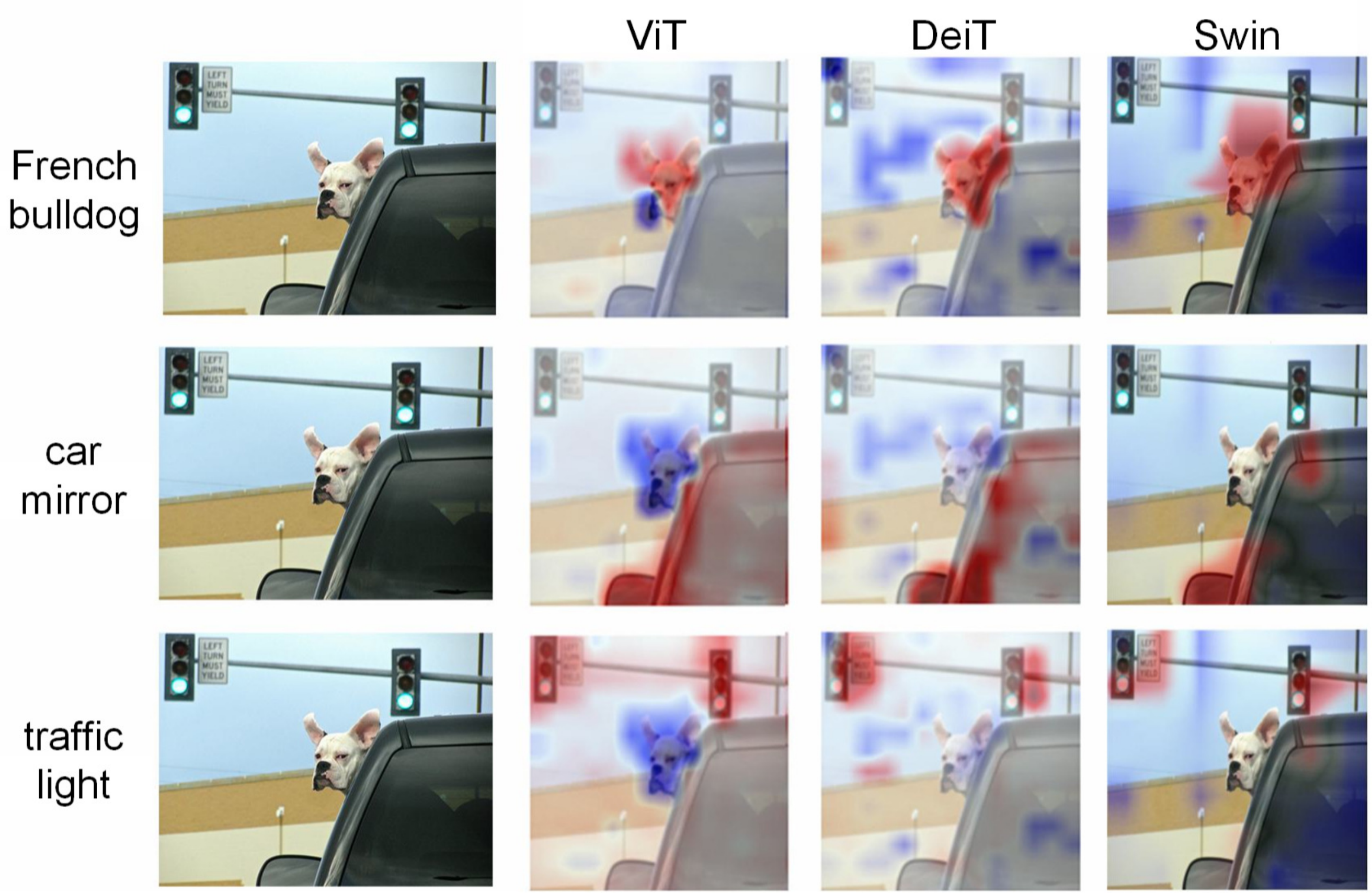}
\caption{BiCAM generalizes to ViT variants, tested on ImageNet. All models adopt the base version. Largely coherent bidirectional attributions are observed.}
\label{fig:architecture_generalization}
\end{figure}

BiCAM can be easily adapted to other architectures with minimal changes. For \textbf{DeiT}~\cite{touvron2021training}, BiCAM focuses on [CLS] token, ignoring the distillation token. For \textbf{Swin}~\cite{liu2021swin}, we aggregate the last two stages with pooled output. \fref{fig:architecture_generalization} shows that BiCAM produces coherent bidirectional attributions across all these architectures.

\section{Conclusion and Discussion}\label{sec:conclusion}

We introduced BiCAM, a bidirectional attribution method that captures both supportive and suppressive evidence in Vision Transformers. By preserving signed contributions, BiCAM provides more complete explanations and improves localization and faithfulness across multiple benchmarks. The proposed PNR metric further demonstrates that bidirectional attributions encode structural signatures that adversarial perturbations systematically disrupt, enabling lightweight adversarial detection without additional training. These findings suggest that bidirectional attribution is a valuable and underexplored dimension for transformer interpretability.

{\bf Limitations.} A central challenge in XAI is the absence of ground-truth attributions; metrics like IoU and faithfulness are indirect proxies, and a formal user study would provide complementary evidence of interpretive value. Additionally, BiCAM relies on gradient-based signals and thus inherits their sensitivity to model smoothness and gradient noise. Moreover, PNR-based adversarial detection has not been compared with dedicated adversarial detectors or tested against adaptive adversaries.

As transformers proliferate in high-stakes applications, BiCAM advances trustworthy AI by revealing the nuanced interplay of signals within black-box decisions. \textbf{Future directions} include extending bidirectional attribution to multi-modal settings; exploring PNR for out-of-distribution detection; and leveraging suppressive patterns for architectural improvements.

\bibliographystyle{splncs04}
\bibliography{references}
\end{document}